\documentclass[final,5p,times,twocolumn]{elsarticle}
\usepackage{graphicx}
\usepackage{epsfig}
\usepackage{amssymb}
\usepackage{amsthm}
\usepackage{multirow}
\usepackage[table]{xcolor}
\usepackage{float}
\usepackage{subcaption}
\usepackage{soul}
\begin{document}

\begin{frontmatter}

\title{Incorporating Pass-Phrase Dependent Background Models for Text-Dependent Speaker Verification}

\author{A. K. Sarkar, Zheng-Hua Tan}
\address[]{Department of Electronic Systems \\
Aalborg University, Denmark \\
 {\small \tt akc@es.aau.dk, zt@es.aau.dk}}

\begin{abstract}
In this paper, we propose  pass-phrase dependent background models (PBMs) for text-dependent (TD) speaker verification (SV) to  integrate the pass-phrase identification process into the conventional TD-SV system, where a PBM is derived from a text-independent background model through adaptation using the utterances of a particular pass-phrase. 
During training, pass-phrase specific target speaker models are derived from the particular PBM using the training data for the respective target model.
While testing, the best PBM is first selected for the test utterance in the maximum likelihood (ML) sense and the selected PBM is then used for the log likelihood ratio (LLR) calculation  with respect to the claimant model. 
The proposed method incorporates the pass-phrase identification step in the LLR calculation, which is not  considered in conventional standalone TD-SV systems. 
The performance of the proposed method is compared  to conventional text-independent background model based TD-SV systems using either Gaussian mixture model (GMM)-universal background model (UBM) or Hidden Markov model (HMM)-UBM or i-vector paradigms.
In addition, we consider two approaches to build PBMs: speaker-independent and speaker-dependent.
We show that the proposed method significantly reduces the error rates of  text-dependent speaker verification  for the non-target types: target-wrong and imposter-wrong while it  maintains comparable TD-SV performance when imposters speak a correct utterance with respect to the conventional system.
Experiments are conducted on the RedDots challenge and the RSR2015 databases that consist of short utterances.  
\end{abstract}

\begin{keyword}
 Pass-phrase dependent  background models (PBMs), GMM-UBM, HMM-UBM, i-vector, text-dependent, speaker verification 
\end{keyword}

\end{frontmatter}

\section{Introduction}
Speaker verification (SV) \cite{DBLP:journals/ejasp/BimbotBFGMMMOPR04,reynold00} is the process of authentication of a person's claimed identity by analyzing his/her speech signal.
 It is a binary pattern recognition problem  where  a SV system makes the decision by calculating  the log-likelihood ratio (LLR)  between the  claimant and background models (also called alternative/negative hypothesis) for the test signal.
 If the LLR value is greater than a pre-defined threshold, the claimant is accepted and otherwise it is rejected.

Speaker verification systems are broadly divided into two categories: text-independent (TI) and text-dependent (TD). 
In TI-SV, speakers are free to speak any sentences, i.e. phrases, during the  enrollment as well as  the test phases. 
It does not impose any constraint that enrollment and test utterances are to be the  same phrase.
However, TD-SV  systems require speakers to speak within  pre-defined sentences, i.e. fixed pass-phrases during the speaker enrollment and test phases. 

In real-life applications, we need a speaker verification system that is accurate on short utterances.
In this regard, TD-SV systems are the ideal choice. 
Since speakers use the same pass-phrase  during both the enrollment and  test phases,  it provides a well matched phonetic content  between the enrollment and test phases. 
Therefore, TD-SV systems are more accurate  compared to their TI-SV counterparts.
Over the last decades, many techniques have been introduced in literature to improve the performance of TD-SV on short utterances.
Examples are deep neural network (DNN) \cite{Nicolas2014_icassp,Nicolas2014,Yuan2015}, i-vector \cite{Nicolas2014,Deka_ieee2011}, hierarchical multi-Layer acoustic model (HiLAM) \cite{reynold00,Larcher2012}, phone-dependent Hidden Markov model (HMM)  \cite{Kajarekar2001, Auckenthaler1999}, domain adaptation \cite{Hagai2014} and phonetic higher level maximum likelihood linear regression (MLLR) super-vector based features \cite{Stolcke05}.
In \cite{Nicolas2014_icassp,Nicolas2014}, phonetic information is  incorporated into an i-vector system by accumulating statistics from speech with respect to a pre-defined phonetic class through an DNN based automatic speech recognition (ASR) system.
In \cite{Yuan2015}, the intermediate output of the DNN layers are used to vectorize characterization  of speech data.
 HiLAM builds a HMM model by concatenating the speech segment-wise adapted models from the Gaussian mixture model- universal background model (GMM-UBM) \cite{reynold00}. 
In domain adaptation \cite{Hagai2014}, the mismatch between the text-independent  and the text-dependent data is reduced  by transforming the text-independent data to better match the text-dependent task  (using the a-priori transcription knowledge of the text-dependent data). 
In conventional HMM based TD-SV systems \cite{Kajarekar2001, Auckenthaler1999}, phoneme (context) dependent speaker models are built using the knowledge of speech transcriptions. 
In \cite{Stolcke05}, a speech signal is represented by a super-vector concatenation of MLLR transformations estimated with respect to a pre-defined phonetic class (e.g. vowel and consonant) using automatic speech recognition (ASR).

All of these techniques need a background model as an alternative/negative hypothesis for TD-SV.
A single \emph{text-independent  background model} (either gender dependent or independent) is commonly  used in literature where target speakers are represented by  models (say in GMM-UBM framework) derived from the background model.
In \cite{Kenny+2014,KennyInterspeech2014,achintya2012_IJST}, a multiple background model concept is proposed to improve the performance of the conventional speaker verification system, by training background models (BMs) based on the  vocal tract length (VTL) characteristic of target speakers as in \cite{achintya2012_IJST} and pass-phrases  as in \cite{Kenny+2014, KennyInterspeech2014} for  text-independent and text-dependent speaker verification, respectively. During enrollment, target speaker models are derived from the BM based on VTL in \cite{achintya2012_IJST} and pass-phrase of target data in \cite{Kenny+2014, KennyInterspeech2014}).
In the test phase, a test utterance is scored against  the claimant and background models specific for the claimant (defined during the enrollment phase) in order to calculate the log-likelihood ratio.
However, this \cite{Kenny+2014, KennyInterspeech2014,achintya2012_IJST}  does not incorporate the pass-phrase identification process to address the  following two non-target types: target-wrong and imposter-wrong. 
 Recently, in \cite{Kinnunen2016}, the authors proposed  a fusion system which combines the score/decision of an utterance verification system  with a conventional SV system to improve the performance of the TD-SV system against target/imposter-wrong non-target trials.
 
In this paper, we propose pass-phrase specific background models (PBMs) for TD-SV to integrate the utterance identification process (without an extra separate system) into the conventional SV system, aiming at rejecting more  non-target types: target-wrong and imposter-wrong  than the conventional SV system while maintaining the performance for the imposter-correct  non-target type. 
In the proposed method,  PBMs are derived from the \emph{text-independent background model} by pooling the training data for a particular pass-phrase from many speakers. 
During enrollment, pass-phrase specific target speaker models are  derived from the particular PBM with MAP adaptation by using the training data for the respective target model.
In the test phase,  the best PBM  is  selected for \emph{a particular test utterance} in the maximum likelihood (ML) sense and  \emph{the best selected PBM} is used as an alternative hypothesis for the log-likelihood ratio calculation with respect to the claimant model, which differs from \cite{Kenny+2014,KennyInterspeech2014,achintya2012_IJST}. 
Furthermore two strategies are considered for building the PBM: one is called \emph{speaker-independent (SI)} and the other  is \emph{speaker-dependent (SD)} again in contrast to \cite{Kenny+2014,KennyInterspeech2014,achintya2012_IJST} where only SI-PBM concept is considered.  
 In the SI-PBM case, PBMs are built  using data from the non-target speakers \emph{who are not participating in the evaluation task}, which reflects the scenario when no priori knowledge about the particular target speaker is given.
It generates a  global set of PBMs  which are common to all target speakers and  used during their training and testing phases.
In the SD-PBM case, PBMs are built by  using the particular  \emph{target speaker training data} together with  the non-target speakers data.
This yields a separate set of PBMs for the particular target speakers that are used during their training and test phases. 
Since SD-PBM incorporates  a priori knowledge about the particular target speaker in the PBMs,  we  called it \emph{speaker-dependent}.
The main salient feature of the proposed method is that it incorporates the utterance identification  in the LLR calculation process in the conventional approach in contrast to  \cite{Kenny+2014,KennyInterspeech2014,achintya2012_IJST}.

We  study the performance of the  proposed PBM based TD-SV system under the  GMM and recently proposed HMM modeling  \cite{achintya2016}  and i-vector paradigms.
Experiments are conducted on the RedDots Challenge \cite{RedDots} and the RSR2015 \cite{RSR2015} databases.
The proposed PBM methods  yield significantly better results than those of the conventional text-independent background model based  TD-SV method for non-target types: target-wrong and imposter-wrong while providing a comparable performance for imposter-correct.

The reminder of this paper is organized as follows: Section \ref{sec:base_tech} describes the GMM-UBM, HMM-UBM and i-vector based TD-SV methods.
Section \ref{sec:propmethod} describes the proposed methods.
Experimental setup  is given in Section \ref{sec:exp_setup}.
Results and discussion are presented in Section \ref{sec:Result_discusiion}.
Finally, the paper is concluded in Section \ref{sec:con}.

\section{Text-dependent speaker verification methods}
In this section, we briefly describe the GMM-UBM, HMM-UBM and i-vector techniques for text-dependent speaker verification, which are considered for comparing the performance of the proposed methods with the existing ones.

\label{sec:base_tech}
\subsection{Gaussian mixture model- universal background model based method}
\begin{figure}[h]
\centering{\includegraphics[height=3.6cm,width=6.5cm]{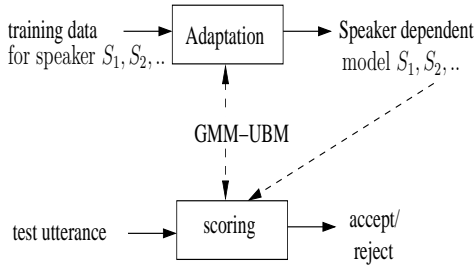}}
\caption{\it GMM-UBM based speaker verification.}
\label{fig:GMM_baseline}
\end{figure}

In this method, a large Gaussian mixture model called GMM-UBM \cite{reynold00} is built using data with different textual contents from non-target speakers.
It represents  a large acoustic space that covers all sorts of  attributes available in the training data.
Then, pass-phrase-specific target speaker models are derived from the text-independent GMM-UBM   with maximum a posteriori (MAP) adaptation using the training data for the particular target model.
In the test phase, a test utterance $X=\{x_1,x_2,\ldots, x_L\}$ is scored against the claimant $\lambda_r$ and  GMM-UBM $\lambda_{UBM}$ for log likelihood ratio calculation,
\begin{eqnarray}
\Lambda(X) =\frac{1}{L} \sum_{t=1}^{L} \big \{ log\; p(x_t|\lambda_r) - log \;p(x_t|\lambda_{UBM}) \big \} 
\end{eqnarray}
where $p(x_t|\lambda_r)$ denotes the likelihood value for a given feature vector $x_t$ with respect to the model $\lambda_r$.
The GMM-UBM  can be either gender-dependent or gender-independent. 
In gender-dependent case, speaker models are derived from the GMM-UBM with respect to their gender.
 It is well established that the gender-dependent GMM-UBM system shows slightly better performance than the gender-independent one.
Fig.\ref{fig:GMM_baseline} illustrates how a  GMM-UBM speaker verification system works.

\subsection{Hidden Markov model - universal background model  based method}
In this method \cite{achintya2016}, an HMM \emph{called `HMM-UBM'} is built using data from many non-target speakers without any speech transcription.
A dummy word (such as  `Hi') is used as the label (\emph{without phonetic level break up}) for all training speech data during the HMM training.
HMM-UBM is trained by pooling all training data together and iteratively updating its parameters using the Baum-Welch re-estimation algorithm.
Since no transcriptions are considered during HMM training, state transition probabilities of the HMM-UBM \cite{Rabiner89} will inherently reflect  the speaker-independent temporal  information available within the  data.
However, this temporal information is not accounted for in the conventional GMM-UBM based TD-SV systems.

\begin{figure}[h]
\centering{\includegraphics[height=2.0cm,width=6.2cm]{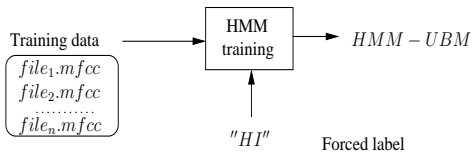}}
\caption{\it Training of un-supervised HMM-UBM with forced dummy label as a transcription to all training data.}
\label{fig:HMM-UBM}
\end{figure}

Similarly to the GMM-UBM TD-SV system,  pass-phrase-specific target speaker models are derived from the text-independent HMM-UBM with MAP adaptation \cite{Gauvain94} using the training data for the particular target  model.
In test, a test utterance is forced-aligned  against the claimant and the HMM-UBM models for the LLR score calculation.
Fig.\ref{fig:HMM-UBM} illustrates the training of  HMM-UBM in an un-supervised manner for text-dependent speaker verification.

It should be noted that the HMM-UBM approach is different from the HiLAM proposed in \cite{Larcher2012} in two ways:
First, HMM-UBM is trained by pooling all training data together using the Baum-Welch re-estimation algorithm, and then target speaker models are derived from the HMM-UBM with MAP using the training data of particular target speaker model.
On the other hand,  HiLAM builds an target specific HMM  by concatenation of speech segment-wise adapted GMM models as \emph{state} (from GMM-UBM) and uniformly assigning  the transition probabilities across the states of HMM.
Secondly, in the HMM-UBM method, both claimant and background models are considered under the same HMM modeling paradigm.
 HiLAM, however, uses HMM for target speaker modeling  and GMM-UBM as a negative  hypothesis during test.

\subsection{i-vector based method}
In this method, a speech utterance is represented by a vector in a low dimensional subspace  in the GMM-UBM super-vector domain called \emph{total variability space} where speaker and channel information is assumed dense.
It is generally expressed as,
\begin{eqnarray}
M = m + Tw \label{eq:ivectr}
\end{eqnarray}
where $w$ is called an i-vector.  $M$, $m$ and $T$ denote the utterance dependent GMM super-vector, the speaker-independent GMM super-vector obtained by concatenating the mean vectors from the GMM-UBM and the total variability space, respectively. 
The following steps are involved during i-vector estimation for a given speech signal $X=\{x_1,x_2,\ldots, x_L\}$ using the  GMM-UBM $\lambda_{UBM}$ and $T$ space.
\begin{itemize}
\item[-] Estimate sufficient statistics,
\begin{eqnarray}
 Pr(c|x_t) & = & \frac{\omega_c p_c(x_t)}{\sum_{j=1}^{C} \omega_j p_j(x_t)} \\
(0^{th}\; order) N_c & =& \sum_{t=1}^{L} Pr(c|x_t,\lambda_{UBM})  \label{eq:first_stat} \\
(1^{st} order)\; F_c & = & \sum_{t=1}^{L} Pr(c|x_t,\lambda_{UBM})x_t 
\end{eqnarray}
 
\item[-] Centralize $F_c$ statistics w.r.t GMM-UBM
\begin{eqnarray}
\tilde{F}_c & = &\sum_{t=1}^{L} Pr(c|x_t,\lambda_{UBM})(x_t-\mu_c) \label{eq:central_sts}
\end{eqnarray}

\item[] - Obtain  i-vector $w$ for $X$ using the statistics,
\begin{eqnarray}
w=(I+T^{'}\Sigma^{-1} N(X)T)^{-1}.T^{'}\Sigma^{-1}\tilde{F}(X) \label{eq:i-vector}
\end{eqnarray}
\end{itemize}
where $c$, $N(X)$ and $\tilde{F(X)}$ represent the $c^{th}$  mixtures of GMM-UBM, the $CF\times CF$ block diagonal matrix, and the $CF\times1$ super-vector obtained by concatenating the first order statistics $\tilde{F}_C$ for the utterance $X$, respectively. $Pr(c|x_t)$ denotes the posteriori alignment of feature vector $x_t$ corresponding to the mixture component $c$. $\omega_c$ and $p_c$ indicate the weight and Gaussian density function of $c^{th}$ mixture component of GMM-UBM, respectively. $\Sigma$ represents a diagonal covariance matrix of dimension $CF\times CF$ estimated during factor analysis training.
$C$ and $F$ are respectively, the number of mixtures in GMM-UBM and the dimension of feature vector.
$\mu_c$ denotes the mean vector corresponding to the $c^{th}$ mixture of GMM-UBM.
$(')$ is the matrix transpose operation. $I$ denotes the identity matrix. 
The total variability space $T$ is trained using data from many non-target  speakers in the expectation-maximization (EM) sense.
For more details about the i-vector see \cite{Deka_ieee2011}.

During enrollment, each target is represented by an average i-vector. The average i-vector is computed over the i-vectors of each speech file available for training the particular target model. An i-vector for a particular speech file is extracted using sufficient statistics with respect to the GMM-UBM,  followed by projection onto the total variability space $T$. 
In the test phase, the score between the  i-vector of the test utterance and the claimant specific is calculated using probabilistic linear discriminate analysis (PLDA). 
PLDA is a generative modeling approach which decomposes the i-vector into several components with a joint factor analysis (JFA) \cite{kenny05}  framework as:
\begin{eqnarray}
w=\mu_w+\Phi y + \Gamma z+\epsilon \label{eq:plda1}
\end{eqnarray}

where $\Phi$ and $\Gamma$ are  matrices denoting the \emph{eigen voice} and \emph{eigen channel} subspaces, respectively. $y$ and $z$ are the speaker and channel factors, respectively, with a priori normal distribution. $\epsilon$ represents the residual noise. $\Phi$, $\Gamma$ and $\epsilon$ are  iteratively updated  during training process by pooling a numbers of  i-vectors per speaker class from many speakers. 
During  test, the score between two i-vectors ($w_1$, $w_2$) is calculated as:
\begin{eqnarray}
score(w_1,w_2)=\log\frac{p(w_1,w_2|\theta_{tar})}{p(w_1,w_1|\theta_{non})} \label{eq:pldascore}
\end{eqnarray}
 hypothesis $\theta_{tar}$ states that $w_1$ and $w_2$ come from the same speaker, and  hypothesis $\theta_{non}$ states that they are from different speakers. For more details about the PLDA based scoring see~\cite{Pierre-interspeech2012,Prince2012,SenoussaouiInterspch2011}.

Prior to PLDA, i-vectors are post-processed for session variability compensation using an iterative conditioning algorithm called \emph{spherical normalization (Sph)} proposed in \cite{Pierre-interspeech2012}.
It has been shown in \cite{Pierre-interspeech2012} that \emph{Sph} improves the speaker verification performance  of PLDA based systems when compared to other conventional approaches. 

\section{Proposed methods}
\label{sec:propmethod}
\subsection{Gaussian mixture model- universal background model based PBM TD-SV  system}
\label{sec:GMM-PBM}
In this case, pass-phrase specific background models  are derived from the text-independent GMM-UBM with MAP  adaptation by pooling the data of the particular pass-phrase of many speakers.
Concerning the real scenarios, we consider two approaches: one is called \emph{speaker-independent} and the other is \emph{speaker-dependent}.
\begin{itemize}
\item {\bf Speaker-independent  PBM (SI-PBM)}: PBMs are built using data from non-target speakers who do not participate in the evaluation task. It  gives  a  global set of PBMs common to all target speakers.
Since this PBM approach does not use any data/information from a particular target speaker, we call it  \emph{speaker-independent}.
\item {\bf Speaker-dependent  PBM (SD-PBM)}: In this case,  each target speaker specific PBMs are built by pooling the pass-phrase specific training data of both the \emph{particular target speaker} and \emph{non-target speakers}. 
This creates PBMs specific for the particular target speaker.
Since SD-PBMs use the training data from a particular target speaker, therefore contains priori information about the particular target speaker and thus call it \emph{speaker-dependent}.
\end{itemize}

After building the PBMs, pass-phrase specific target speaker models  are derived from the particular PBM with MAP adaptation by using his/her training data for the particular target model. \emph{Algorithm 1} describes  the enrollment phase of the target speaker model using PBMs.\\
\hrule
\vspace*{+0.1cm}
{\bf Algorithm 1}: Enrollment phase
\vspace*{+0.1cm}
\hrule
\begin{itemize}
\item[] {\emph{Initial}: } load the PBMs 
\item[] {\bf Step1:} Read \emph{all} the enrollment data $X_r$ of pass-phrase specific target model, $r$  
\item[] {\bf Step2:} Choose the PBM  as per enrollment data
\item[] {\bf Step3:} Derive the target  model $\lambda_r$  from the chosen PBM with MAP adaptation using the enrollment data $X_r$
\item[] {\bf Step4:} Repeat the Step 1 to 3 for all target models
\end{itemize}
\vspace*{+0.1cm}
\hrule
\vspace*{+0.3cm}

In the test phase, \emph{the best PBM} for the particular test utterance $Y=\{y_1,y_2,\ldots, y_L\}$ is first selected from  the PBMs obtained in training phase in the ML sense. Then, the selected  PBM  is used as an alternative hypothesis for the log likelihood ratio calculation with respect to  the claimant model, $\lambda_r$. 
\emph{Algorithm 2} presents the test phase of the proposed method.\\
\hrule
\vspace*{+0.1cm}
{\bf Algorithm 2}: Test phase
\vspace*{+0.1cm}
\hrule
\begin{itemize}
\item[] {\bf Step1:} Load the feature vectors of a test utterance $Y=\{y_1,y_2,\ldots, y_L\}$, PBMs and the target model $\lambda_r$ for the claimant, $r$
\item[] {\bf Step2:} Find the best PBM in the ML sense for the test utterance,
\begin{eqnarray}
        \hat{i}_{PBM}= \mbox{arg }\max_{i} p(Y |\lambda_{PBM_i}) \label{ML_alpha2}
\end{eqnarray}
\item[] {\bf Step3:} Calculate the log likelihood ratio as,
\begin{eqnarray}
\Lambda(Y) =\frac{1}{L} \sum_{t=1}^{L} \big \{ log\; p(y_t|\lambda_r) - log \;p(y_t|\lambda_{\hat{i}_{PBM}}) \big \} \label{eq:LLR_eq}
\end{eqnarray}
\end{itemize}
\vspace*{+0.1cm}
\hrule
\vspace*{+0.3cm}

\begin{figure}[!h]
\centering{\includegraphics[height=6.5cm,width=8.2cm]{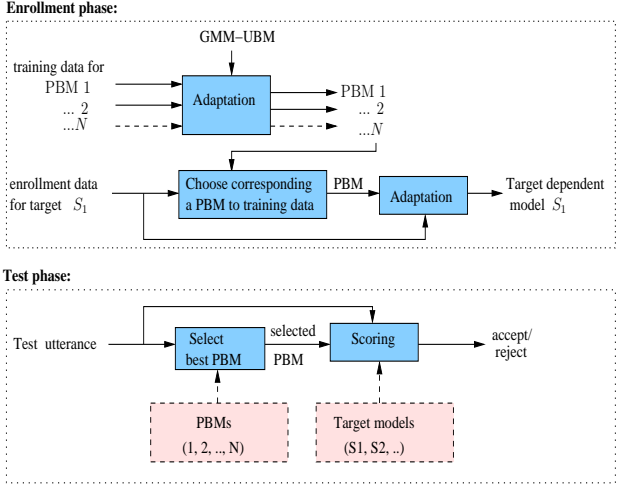}}
\caption{\it Pass-phrase dependent background models based  text-dependent speaker verification system in GMM-UBM paradigm.}
\label{fig:GMM-PBM}
\end{figure}

Based on speaker independent/dependent PBMs, it yields  two sub-systems:  \emph{GMM-UBM speaker-independent PBM (GMM-UBM-SI-PBM)} and  \emph{GMM-UBM speaker-dependent PBM (GMM-UBM-SD-PBM)} TD-SV systems.
Fig.\ref{fig:GMM-PBM} shows the block diagram of the proposed PBM based text-dependent speaker verification system in the GMM-UBM paradigm.

\subsection{Hidden Markov model- universal background model based PBM TD-SV system}
\label{sec:HMM_PBM}
This system is similar to the \emph{GMM-UBM based PBM TD-SV  system}. 
The main difference is that HMM-UBM is considered instead of GMM-UBM.

As we also consider both speaker independent and dependent training strategies for building the PBMs, we get two sub-systems called \emph{HMM-UBM speaker-independent PBM (HMM-UBM-SI-PBM)}  and \emph{HMM-UBM speaker-dependent PBM (HMM-UBM-SD-PBM )} TD-SV systems.

\subsection{i-vector based PBM TD-SV system}
\label{sec:ivector_PBM}
We further investigate the GMM-UBM PBM based TD-SV system in i-vector paradigm.
In the enrollment phase, zero and first order statistics are estimated with respect to the \emph{particular PBM} based on the \emph{target model of specific pass-phrase} for \emph{all training speech files} that individually belongs to the respective target model.
Then, statistics are  projected onto  the \emph{total variability space}, $T$  to get an \emph{i-vector for each speech file}.
After that, the target is represented by an average i-vector computed over the speech file-wise i-vector.
Before projecting  the first-order statistics onto the  $T$ space, they are centralized with respect to the \emph{text-independent GMM-UBM} i.e. both the proposed method and conventional system use the same $T$ space. \emph{Algorithm 3} describes the enrollment phase of the target speakers in the PBM based i-vector system.
It is also noted that  PBMs are derived from the  GMM-UBM with MAP only updating the Gaussian mean vectors and hence  all the PBMs and GMM-UBM  share the same Gaussian weight and covariance matrix (except mean) parameters across the Gaussian components.
\\
\hrule
\vspace*{+0.1cm}
{\bf Algorithm 3: Enrollment phase}
\vspace*{+0.1cm}
\hrule
\begin{itemize}
\item[] {\bf Initial:} Load GMM-UBM, $T$ matrix and PBMs
\item[] {\bf Step1:} Read \emph{all} enrollment data $X_r$ for the  target model, $r$ (for a particular pass-phrase)  
\item[] {\bf Step2:} Choose the PBM (e.g. $\lambda_{PBM}$) as per pass-phrase of enrollment data
\item[] {\bf Step3:}  Estimate sufficient statistics for $a^{th}$ speech file $\epsilon \;X_r$, say $X^{a}_r=\{x_1,x_2,\ldots,x_L\}$  w.r.t $\lambda_{PBM}$ model,
\begin{itemize}
\item[]
\begin{eqnarray}
(0^{th}\; order) N_c & =& \sum_{t=1}^{L} Pr(c|x_t,\lambda_{PBM})  \label{eq:first_stat} \\
(1^{st} order)\; F_c & = & \sum_{t=1}^{L} Pr(c|x_t,\lambda_{PBM})x_t 
\end{eqnarray}
\item[] - Centralized $F_c$ statistics w.r.t GMM-UBM
\begin{eqnarray}
\tilde{F}_c & = &\sum_{t=1}^{L} Pr(c|x_t,\lambda_{PBM})(x_t-\mu_c) \label{eq:central_sts}
\end{eqnarray}

\item[] - Obtained  i-vector $w^{a}_r$ using the statistics for $X^{a}_r$ as,
\begin{eqnarray}
w^{a}_r=(I+T^{'}\Sigma^{-1} N(X^a_r)T)^{-1}.T^{'}\Sigma^{-1}\tilde{F}(X^a_r) \label{eq:i-vector}
\end{eqnarray}
\vspace*{+0.1cm}
\end{itemize}

\item[] {\bf Step4:} Repeat the Step 3 for ($\#$) number of speech files 
\item[] {\bf Step5:} Compute average i-vector for target model, $r$
{\footnotesize
\begin{eqnarray}
 w_r = \frac{1}{\#}\sum_{a=1}^{\#} w_r^a
\end{eqnarray}
}
\item[] {\bf Step6:} Repeat the Step 1 to 5 fo all target models

\end{itemize}

\hrule
\vspace*{+0.2cm}

In the test phase, \emph{the best PBM} is first selected for the test utterance in the ML sense (similarly to Eqn(\ref{ML_alpha2})) and following  sufficient statistics are computed with respect to the selected PBM (similarly to Eqn(\ref{eq:first_stat}-\ref{eq:central_sts})). 
Finally, statistics are projected onto the $T$ space for i-vector extraction (Eqn.\ref{eq:i-vector}) for the test utterance.
The score between the two i-vectors: claimant specific and the test utterance is calculated with PLDA.

Based on the PBM training strategies, it gives us two subsystems: i-vector speaker-independent PBM (i-vector-SI-PBM) and i-vector speaker-dependent PBM (i-vector-SD-PBM) for TD-SV, respectively.

It is  important to note that $T$ space, i-vectors for training PLDA and EFR are extracted using the statistics with respect to the text-independent GMM-UBM i.e.  i-vector baseline system represented in Sec.\ref{sec:base_tech}, and also reused in the proposed method.
It gives an identical setup for comparison the proposed method with the baseline system.

\section{Experimental setup}
\label{sec:exp_setup}
Experiments are performed on male speakers in two databases: RedDots challenge (task \emph{m-part-01}) and RSR2015  (evaluation set of part1 i.e. \emph{3sess-pwd\_eval\_m} task) as per protocols in \cite{RedDots} and \cite{RSR2015}, respectively.
The respective task in each database  consists of the speakers data recorded on common pass-phrases (i.e. sentences) over many sessions for text-dependent speaker verification.
There are three recording sessions to train the particular pass-phrase-wise target speaker model.
The utterances are of very short duration on an average of 2-3s per speech signal. 
Test trials are divided into three types of non-target  for system performance evaluation:
 \begin{itemize}
\item {\bf target wrong:}  when a target speaker speaks a wrong  sentence i.e. a different pass-phrase,  in the testing phase as compared to their enrollment phase.
\item {\bf imposter correct:} imposter speaks a correct sentence, i.e. the same pass-phrase as that of the target enrollment sessions.
\item {\bf imposter wrong:} imposter speaks a wrong sentence i.e. a different pass-phrase from target enrollment phrase.
\end{itemize}

In the RSR2015 database, there are 1708 target models for training and evaluation.  Data from $50$ development speakers (disjoint from the evaluation set) are used for training SI-PBMs. In case of the RedDots database, a disjoint set of nine speakers' data (excluded from the evaluation) are used for SI-PBMs (approximately $148$ files per pass-phrase) and the remaining speakers are considered for the evaluation \cite{Kinnunen2016}.
This gives  $248$ target models for training and evaluation trials  from $40$ different speakers. 
Table \ref{table:trial_info} shows the number of trials available for the system evaluation on different types of non-target scenarios on the respective databases.
\begin{table}[h!]
\caption{\it Number of trials for system evaluation on different databases.}
\begin{center}
\begin{tabular}{l|l|l|l|l}\cline{1-5}
Database &No of & \multicolumn{3}{c}{No of trials in Non-target type} \\ \cline{3-5}
          & True      & Target  & Imposter & Imposter \\
          &  trials     &-wrong   &-correct & -wrong\\ \hline \hline
RedDots   & 2119        & 19071        & 62008            & 557882  \\  
RSR2015   &10244        & 297076       & 573664           & 8318132  \\ \hline
\end{tabular}
\end{center}
\label{table:trial_info}
\vspace*{-0.2cm}
\end{table}

For SV systems on RSR2015, text-independent GMM-UBM and HMM-UBM are trained by pooling data consisting of various textual contents from the TIMIT database across $438$ male non-target speakers (approximately 4380 utterances).
The GMM/HMM-UBM training data are reused for the training of \emph{total variability space}. 
For PLDA and Sph, training data used for GMM/HMM-UBM and SI-PBMs ($50$ development speakers who are disjoint from the evaluation set) are used.
As  part 1 of RSR2015 database contains  recordings of speaker data over 30 common pass-phrases in 9 sessions. This gives 30 PBMs (approximately 450 files per pass-phrase) in the proposed methods.  
In PLDA training, utterances having the same pass-phrase of a particular speaker (of $50$ development speakers from the RSR2015 database) are treated as belonging to the individual speaker class. 
This gives a total of $1938$ classes in PLDA ($438$ from TIMIT and $1500$ from the development set of RSR2015).

For SV systems on  RedDots, GMM-UBM and HMM-UBM  are trained using data of various textual contents from the RSR2015 database \cite{RSR2015} over $157$ male non-target speakers (approximately $42000$ utterances). Since the data of the speakers in the \emph{m-part-01 task of RedDots database} were recorded over $10$ common pass-phrases, it yields  $10$ PBMs  in the proposed PBM based systems.
The GMM/HMM-UBM training data are reused for the training of \emph{total variability space}, PLDA and Sph algorithm. 
In PLDA training, utterances having the same pass-phrase of a particular speaker are treated as belonging to the individual speaker class. 
This gives a total of $157*30=4710$ speakers (classes) in PLDA (each class having on average $9$ examples).

 GMM based systems consist of GMM-UBM with $512$ Gaussian mixtures. In case of HMM-UBM, 14 states and different numbers of Gaussians per state are considered as per \cite{achintya2016} and inspired from \cite{Zheng_hua2010}. 
 A left to right modeling concept is followed in the HMM-UBM modeling.
 HTK toolkit \cite{htkbook} and ALIZE toolkit \cite{alizetoolkit} are used for implementing the HMM-UBM and GMM-UBM based systems, respectively.
Both GMM- and HMM-UBM have diagonal covariance matrices. During the construction of GMM-UBM PBMs  only Gaussian mean vectors are updated with  MAP adaptation of GMM-UBM. For HMM-UBM PBMs, both Gaussian mean vectors and transition parameters are updated during MAP adaptation of HMM-UBM. When  enrolling a target, only Gaussian mean vectors of the GMM/HMM-UBM/PBMs are updated during MAP adaptation.  Three iterations are used in MAP with a value of relevance factor $10$ in all cases.

In i-vector based systems, channel and speaker factors in PLDA are kept equal to the dimension of i-vector. In Sph, two iterations are followed. We consider i-vector of $400$ dimensions. 
An i-vector is extracted for each single utterance. During enrollment, each target is represented by an average i-vector computed over their respective training example-wise i-vector.
It is noted that i-vectors for training both Sph and PLDA are  extracted from the GMM-UBM rather than from PBM. In other words,  total variability space,  PLDA and Sph parts are the same for both conventional and the proposed method. The BOB toolkit \cite{bob2012} is used for implementing the i-vector system.

For cepstral analysis, $57$ dimensional MFCC \cite{Davis80} ($19$ static+$\Delta$+$\Delta\Delta$) feature vectors are extracted from speech signals using $20$ms hamming window  with $10$ms overlap of adjacent frames.
 RASTA \cite{Hermanksy94} filtering is applied on the features.
 An energy based voice activity detection (VAD) is used to discard the low energy frames.
High energy frames are normalized to fit zero-mean and unity variance at the utterance level.
System performances are evaluated in terms of equal error rate (EER) and minimum detection cost function (MinDCF) \cite{DET97}.

\section{Results and Discussion}
\label{sec:Result_discusiion}
\begin{table*}[t]
\caption{\it Comparison performance of the proposed PBM methods with the baseline system for text-dependent speaker verification on different databases with GMM-UBM paradigm.}
\begin{subtable}{20.0cm}
\caption{ On task \emph{m-part-01} of {\bf RedDots}} 
\begin{center}
\begin{tabular}{llll}\cline{1-4}
System                    & \multicolumn{3}{c}{Non-target type [\%EER/(MinDCF$\times$ 100)]} \\
                          & target-wrong      & imposter-correct & imposter-wrong \\ \hline \hline
1. Baseline GMM-UBM       &4.64/2.396        &  2.60/1.223      & 0.66/0.245            \\
2. Prop. GMM-UBM-SI-PBM   & {\bf 0.76/0.511 }  & {\bf 2.42/1.082} & {\bf  0.33/0.103}   \\
3. Prop. GMM-UBM-SD-PBM   &{\bf 0.33/0.224}    & 2.83/{\bf 1.060} & {\bf 0.33/0.145 }  \\ \hline
\end{tabular}
\end{center}
\label{table:GMM-res-reddots}
\end{subtable}
\\[2.8ex]
\begin{subtable}{20.0cm}
\caption{ On male evaluation set of part1 (\emph{3sess-pwd\_eval\_m task})  of {\bf RSR2015}}      
\begin{center}
\begin{tabular}{llll}\cline{1-4}
System                    & \multicolumn{3}{c}{Non-target type [\%EER/(MinDCF$\times$ 100)]} \\
                          & target-wrong        & imposter-correct & imposter-wrong \\ \hline \hline
1. Baseline GMM-UBM       & 1.07/0.530          & 1.46/0.701        & 0.14/0.060            \\
2. Prop. GMM-UBM-SI-PBM   & {\bf 0.009/0.008 } & {\bf 0.80/0.353}  & {\bf  0.004/0.001}   \\
3. Prop. GMM-UBM-SD-PBM   & {\bf 0.009/0.002}  & {\bf 0.79/0.339} & {\bf 0.004/0.001}    \\ \hline
\end{tabular}
\end{center}
\label{table:GMM-res_rsr}
\end{subtable}
\label{table:GMM_res}
\end{table*}

\subsection{Systems based on Gaussian mixture model}
\label{sec:GMM_domain}
Tables \ref{table:GMM-res-reddots} and  \ref{table:GMM-res_rsr} compare the performance of the proposed PBM methods with the baseline system for text-dependent speaker verification on the GMM-UBM paradigm. 
It can be seen  that the proposed PBM techniques give significantly lower  EER and MinDCF values for the non-target types: target-wrong and imposter-wrong than the baseline system on both databases. 
At the same time, the performance of the proposed methods for the imposter-correct type  is comparable/closer to the baseline. This shows the effectiveness of the proposed PBM methods which incorporates the pass-phrase identification process in LLR during the test phase. To provide inside information about the significant performance improvement for target-wrong and imposter-wrong, we calculate the \emph{LLR score difference} between the baseline and the proposed SI- and SD-PBM based systems i.e. $Y(x)= LLR_{baseline}(x)$ $-$  $LLR_{PBM}(x)$ where $x$ denotes a trial for the target-wrong and imposter-wrong trials.
It shows that 99.67\% and 99.86\% of the target-wrong and imposter-wrong trials  show lower LLR score values, respectively  for the SI-PBM and SD-PBM systems  than the baseline system  in the RedDots database. 
Similar observation is also seen on the RSR2015 database.  
This further demonstrates  that the proposed PBM based methods are able to reject target-wrong and imposter-wrong non-target types more than the conventional TD-SV method.

The SD-PBM system shows slightly lower error rates  than  SI-PBM for the non-target type: target-wrong.
This is expected due to the fact that SD-PBMs contain the priori information about the target speakers and hence it would give a lower  pass-phrase identification error rate during test.
It is also well known in automatic speech recognition (ASR) that a SD system  gives higher accuracies than the SI counterpart. 

Table \ref{table:IER} shows the pass-phrase identification accuracies  on the evaluation data for different PBM based systems on both databases.
From  Table \ref{table:IER}, we can see that SD-PBM systems show higher pass-phrase identification accuracies than SI-PBM systems on the evaluation set, which is also reflected by the speaker verification performance in the Tables \ref{table:GMM-res-reddots} (SD-PBM shows lower error rates for non-target type target-wrong than SI-PBM).
On RSR2015 (Table \ref{table:GMM-res_rsr}), both SI- and SD-PBM systems show equal (and saturated) value of pass-phrase identification accuracy.
Both PBM systems also show very close speaker verification performance to each other.
Table \ref{table:IER} further shows that the system gives an approximately 4\%  pass-phrase identification error rate on the evaluation data of the RedDots database.
However, the proposed method still shows significantly lower error rates for non-target types: the target/imposter-wrong cases while maintaining a  performance for the imposter-correct non-target type comparable  to the baseline SV system.

\begin{table}[h!]
\caption{\it Pass-phrase identification accuracy (\%) of GMM-UBM-SI/SD PBM systems on evaluation data over different databases.}
\begin{center}
\begin{tabular}{llll}\cline{1-3}
Database  & PBM & evaluation set \\  \hline \hline
RedDots   & SI  & 95.97   \\
          & SD  & 96.30   \\ \\
RSR2015   & SI  & 99.94  \\
          & SD  & 99.94 \\ \hline
\end{tabular}
\end{center}
\label{table:IER}
\vspace*{-0.2cm}
\end{table}

In case of the non-target type: imposter-correct  in the RedDots database, the SD-PBM system shows slightly higher speaker verification error rates  than the  SI-PBM and the baseline.
The reason for higher error rates in the SD-PBM could be due to the fact that  both SI-PBM and baseline systems do not use the target speaker data during training the PBM and GMM-UBM, respectively.
 Hence, SI-PBM and GMM-UBM  theoretically satisfy the fully non-target  hypothesis than the SD-PBM  with respect to the targets.
Therefore, the false rejection rate  is expected to be increased on a certain operating region of the SD-PBM system  with respect to the  SI-PBM and the baseline.
This impacts on the equal error rate.
However, there is always a trade-off between the false acceptance and false rejection rate based on the intended application.
\begin{table*}[!ht]
\caption{\it Comparison performance of the proposed PBM methods with baseline system for text-dependent speaker verification on different databases on  HMM-UBM paradigm.}
\begin{subtable}{20.0cm}
\caption{ On task m-part-01 of {\bf RedDots}}
\begin{center}
\begin{tabular}{lllll}\cline{1-5}
System                   & No of states/        & \multicolumn{3}{c}{Non-target type [\%EER/(MinDCF$\times$ 100)]} \\
                         & mixtures             &                &                    & \\
                         & per state            & target-wrong      & imposter-correct & imposter-wrong \\ \hline \hline
1. Baseline HMM-UBM      & 14/8                 & 4.23/1.925        & 4.01/2.164       & {\bf  1.03}/0.425           \\
                         & 14/16                & {\bf 4.17/1.860}  & 3.68/1.950       & 1.22/{\bf 0.360} \\
                         & 14/32                & 4.61/1.969        & 3.58/1.696       & 1.41/0.466 \\
                         & 14/64                & 4.71/2.105        & {\bf 3.33/1.550} & 1.55/0.500 \\ \\
2. Prop. HMM-UBM-SI-PBM  & 14/8                 & 2.31/1.206        & 3.01/1.275       & 0.89/0.376    \\
                         & 14/16                & 1.74/1.003        & 2.97/1.208       & 0.84/0.320 \\
                         & 14/32                & 1.34/0.791        & {\bf 2.63/1.064} & 0.61/0.285 \\
                         & 14/64                & {\bf 1.09/0.704}  &  2.73/1.017      & {\bf 0.47/0.237} \\ \\
3. Prop. HMM-UBM-SD-PBM  & 14/8                 & 1.93/1.024        & {\bf 3.26/1.351} & 1.03/{\bf 0.397}   \\ 
                         & 14/16                & 1.51/0.845        & 3.91/1.458       & 0.87/0.412  \\ 
                         & 14/32                &0.84/0.632         & 4.90/1.494       & 0.80/0.418 \\
                         & 14/64                & {\bf 0.42/0.407}  & 5.56/1.533       & {\bf 0.64}/0.462  \\ \hline
\end{tabular}
\end{center}
\label{table:HMM-res-reddots}
\end{subtable}
\\[2.8ex]
\begin{subtable}{20.0cm}
\caption{ On male evaluation set of part1 (\emph{3sess-pwd\_eval\_m task}) of {\bf RSR2015}}
\begin{center}
\begin{tabular}{lllll}\cline{1-5}
System                   & No of states/        & \multicolumn{3}{c}{Non-target type [\%EER/(MinDCF$\times$ 100)]} \\
                         & mixtures             &                &                    & \\
                         & per state            & target-wrong   & imposter-correct & imposter-wrong \\ \hline \hline
1. Baseline HMM-UBM      & 14/8                 & 0.53/0.270     & 1.84/0.929       & 0.12/0.038           \\
                         & 14/16                & 0.53/0.239     & 1.53/0.759       & 0.14/0.050  \\
                         & 14/32                & 0.51/0.201     & 1.30/0.656       & 0.08/0.040 \\
                         & 14/64                & {\bf 0.44/0.197}& {\bf 1.16/0.537}& {\bf 0.08/0.027} \\ \\
2. Prop. HMM-UBM-SI-PBM  & 14/8                 & 0.20/0.095     & 0.78/0.355       & 0.05/0.025    \\
                         & 14/16                & 0.10/0.066     &  0.77/0.320      & 0.03/0.0128 \\
                         & 14/32                & 0.07/0.042     & 0.71/0.296       & 0.028/0.015 \\
                         & 14/64                & {\bf 0.02/0.020}& {\bf 0.68/0.287}& {\bf 0.009/0.006}\\ \\
3. Prop. HMM-UBM-SD-PBM  & 14/8                 & 0.20/0.090         & 0.80/0.358        & 0.05/0.026   \\
                         & 14/16                & 0.08/0.055        & 0.82/0.323       & 0.03/0.013  \\ 
                         & 14/32                & 0.06/0.0351        & {\bf 0.79/0.313}       & 0.03/0.018   \\
                         & 14/64                & {\bf 0.01/0.012}        & 0.81/0.337       & {\bf 0.009/0.007} \\ \hline
\end{tabular}
\end{center}
\label{table:HMM-res_rsr}
\end{subtable}
\label{table:HMM-res}
\vspace*{-0.1cm}
\end{table*}

\subsection{Systems based on Hidden Markov model}
Tables \ref{table:HMM-res-reddots} and \ref{table:HMM-res_rsr} compare the performance of the proposed PBM method with the baseline system on the HMM paradigm  for different numbers of mixtures per state in HMM-UBMs. 
Similarly to the GMM-UBM based PBM systems, we can observe from  Tables \ref{table:HMM-res-reddots} -\ref{table:HMM-res_rsr} that the PBM system shows lower error rates compared to  the baseline for non-target types: target-wrong and imposter-wrong in both databases.
For imposter-correct  in the RedDots database, SI-PBM systems  show better or comparable performance to the baseline over various numbers of mixtures in the HMM-UBM state. However, SV error rates of the SD-PBM systems for imposter-correct increase when the number of Gaussian components per state in the HMM-UBM increases  with compared to the SI-PBM and baseline systems. 
This could be due to the same reason as explained in section \ref{sec:GMM_domain} (for the GMM-UBM-SD-PBM system) i.e. SD-PBM uses target data during its training phase. Hence, it does not fully satisfies the non-target hypothesis with respect to the target model in contrast to SI-PBM and HMM-UBM.
It is also noted that pass-phrase identification accuracies of the HMM-UBM based PBM systems are similar to the systems in GMM-UBM paradigm, which is not shown in the paper.
\begin{table*}[!ht]
\caption{\it Comparison performance of the proposed PBM methods with baseline system for text-dependent speaker verification on different databases in  i-vector  paradigm.}
\begin{subtable}{20.0cm}
\caption{ On task m-part-01 of {\bf RedDots}}
\begin{center}
\begin{tabular}{llll}\cline{1-4}
System                    & \multicolumn{3}{c}{Non-target type [\%EER/(MinDCF$\times$ 100)]} \\
                          & target-wrong         & imposter-correct & imposter-wrong \\ \hline \hline
1. Baseline i-vector         & 5.60/2.691        & {\bf 4.15/1.675}  & 1.03/0.470           \\
2. Prop. i-vector-SI-PBM     & {\bf 3.16/1.495}  & 4.38/1.778       & {\bf 0.66/0.236}    \\
3. Prop. i-vector-SD-PBM     & {\bf 3.14/1.386}  & 4.50/1.695       & {\bf 0.56/0.236}   \\ \hline
\end{tabular}
\end{center}
\label{table:ivector-res-reddots}
\end{subtable}
\\[2.8ex]
\begin{subtable}{20.0cm}
\caption{ On male evaluation set of part1 (\emph{3sess-pwd\_eval\_m task}) of {\bf RSR2015}}
\begin{center}
\begin{tabular}{llll}\cline{1-4}
System                    & \multicolumn{3}{c}{Non-target type [\%EER/(MinDCF$\times$ 100)]} \\
                          & target-wrong      & imposter-correct    & imposter-wrong \\ \hline \hline
1. Baseline i-vector         & 2.19/1.077        & 3.08/1.499       & 0.29/0.141           \\
2. Prop. i-vector-SI-PBM     & {\bf 0.56/0.278}  & {\bf 2.86/1.494} & {\bf 0.10/0.0414}    \\
3. Prop. i-vector-SD-PBM     & {\bf 0.55/0.260}  & {\bf 2.81/1.445} & {\bf  0.10/0.039}   \\ \hline
\end{tabular}
\end{center}
\label{table:ivector-res_rsr}
\end{subtable}
\label{table:ivector-res}
\vspace*{-0.2cm}
\end{table*}

\subsection{Systems based on i-vector}
Tables \ref{table:ivector-res-reddots} and \ref{table:ivector-res_rsr} show the performance of the proposed PBM methods incorporated into an i-vector paradigm and that of an i-vector baseline for text dependent speaker verification.
 From Tables \ref{table:ivector-res-reddots} and \ref{table:ivector-res_rsr}, it can be observed that the proposed PBM methods on the i-vector framework show much lower error rates for  the target-wrong and imposter-wrong; and comparable error rates for imposter-correct with compared to the baseline. 
This again indicates the usefulness of the proposed PBM methods of  incorporating the pass-phrase identification process. 
Now if we  compare the PBM based system on the i-vector paradigm with  GMM and HMM, it can be observed from Tables \ref{table:GMM_res}, \ref{table:HMM-res} and \ref{table:ivector-res} that the PBM system based on GMM and HMM more significantly reduces the error rate  for  target-wrong and imposter-wrong. 
This could be due to the nature of very short duration utterances in both the RedDots and RSR2015 databases, and short utterances are challenging for the i-vector paradigm. 
For the non-target imposter-correct type in RedDots database, the error rates of the SI-PBM on the i-vector framework are marginally higher compared to the method in GMM and HMM paradigms.
It could be due to the fact that a selection (i.e. switching) of the PBM during the test phase  introduces another variability in the i-vector system, i.e. the known fact that i-vector based SV systems are very sensible to the session variability.

\section{Conclusion and future work}
\label{sec:con}
In this paper, we proposed  pass-phrase dependent background models (PBMs) for text-dependent speaker verification (TD-SV) to integrate the pass-phrase identification process into conventional TD-SV systems, by deriving PBMs from the text-independent background model with adaptation using the data for a particular pass-phrase across the speakers.
During training, a pass-phrase specific target speaker model is derived from the corresponding PBM using his/her training data of the particular  pass-phrase.
During test, the best PBM for the particular test utterance is selected in the maximum likelihood sense and used for a log likelihood ratio score calculation with respect to the claimant.
The effectiveness of the  proposed techniques were compared with the conventional text-dependent speaker verification systems under GMM-UBM and HMM-UBM paradigms.
We further showed  the proposed concept can be incorporated into an i-vector based TD-SV system.
Experimental results were demonstrated on the recent RedDots challenge and RSR2015 databases.
We showed that incorporation of pass-phase identification in the test phase significantly reduces the speaker verification error rates, especially for the non-target types: target-wrong and imposter-wrong while giving better or comparable performances for the imposter-correct case.
Future work includes better techniques for incorporating PBM into the i-vector paradigm  in order to further improve the speaker verification performance.

\section*{Acknowledgments}
This work is partly supported by the iSocioBot project, funded by
the Danish Council for Independent Research - Technology and Production
Sciences (\#1335-00162) and OCTAVE Project (\#647850), funded
by the Research European Agency (REA) of the European Commission,
in its framework programme Horizon 2020. The views expressed in this
paper are those of the authors and do not engage any official position of
the European Commission.

\bibliographystyle{elsarticle-num}
\bibliography{References}

 \end{document}